\pdfoutput=1

\documentclass[11pt]{article}

\usepackage{acl}

\usepackage{times}
\usepackage{latexsym}

\usepackage[T1]{fontenc}

\usepackage[utf8]{inputenc}

\usepackage{microtype}

\usepackage{linguex}
\usepackage{amsmath}
\usepackage{subcaption}
\usepackage{graphicx}
\usepackage{booktabs}
\usepackage{adjustbox}
\usepackage{tikz-qtree}
\tikzset{every tree node/.style={align=center, anchor=north}}
\usetikzlibrary{positioning}
\usepackage[dvipsnames]{xcolor}

\newcommand{\zero}[1]{#1$^{0}$}
\newcommand{\Prime}[1]{#1${'}$}

\title{SPAWNing Structural Priming Predictions from a Cognitively Motivated Parser 
}

\author{Grusha Prasad \\
  Colgate University \\
  \texttt{gprasad@colgate.edu} \\\And
  Tal Linzen \\
  New York University \\
  \texttt{linzen@nyu.edu} \\}

\begin{document}
\maketitle
\begin{abstract}
Structural priming is a widely used psycholinguistic paradigm to study human sentence representations. In this work we introduce SPAWN, a cognitively motivated parser that can generate quantitative priming predictions from contemporary theories in syntax which assume a lexicalized grammar. By generating and testing priming predictions from competing theoretical accounts, we can infer which assumptions from syntactic theory are useful for characterizing the representations humans build when processing sentences. As a case study, we use SPAWN to generate priming predictions from two theories (Whiz-Deletion and Participial-Phase) which make different assumptions about the structure of English relative clauses. By modulating the reanalysis mechanism that the parser uses and strength of the parser's prior knowledge, we generated nine sets of predictions from each of the two theories. 
Then, we tested these predictions using a novel web-based comprehension-to-production priming paradigm. We found that while the some of the predictions from the Participial-Phase theory 
aligned with human behavior, none of the predictions from the the Whiz-Deletion theory did, thus suggesting that the Participial-Phase theory might better characterize human relative clause representations. 

\end{abstract}

\section{Introduction}

\setlength{\Exlabelwidth}{0.25em}
\setlength{\SubExleftmargin}{1.3em}

Structural priming \citep{branigan2017} is a widely used paradigm in psycholinguistics to study the structural representations that people construct when processing sentences. In this paradigm, researchers measure the extent to which the production or processing of \textit{target} sentences is facilitated (or \textit{primed}) by preceding \textit{prime} sentences, and then use the pattern of priming behavior to draw inferences about the representations people construct. For example, consider a \textit{target sentence} like \ref{ex:po1}. 
\vspace{-0.25em}
\ex. \label{ex:po1} The boy threw the ball to the dog. \vspace{-0.5em}

Prior work \cite{branigan1995} found that targets like \ref{ex:po1} were produced more often, and were processed more rapidly, when they were preceded by primes like \ref{ex:po2}, that have the same structure, than when they were preceded by primes like \ref{ex:do2}, which, while describing the same transfer event as  \ref{ex:po2}, have a different structure. 

\vspace{-0.25em}
\ex. \label{ex:po2} The lawyer sent the letter to the client. \vspace{-0.5em}

\ex. \label{ex:do2} The lawyer sent the client the letter.
\vspace{-0.5em}

From this result, \citeauthor{branigan1995} inferred that participants' mental representation of \ref{ex:po1} is more similar to that of \ref{ex:po2} than of \ref{ex:do2}. 

\citet{branigan2017} propose that by carefully studying which sentences prime each other we can build a theory of human structural representations. Building such a theory requires us to generate hypotheses about the particular prime-target pairs that would be most informative to compare. 
Insights from theoretical syntax, a field that has spent decades studying the structure of sentences, can help constrain this hypothesis space \cite{gaston2017}: if two theories generate different priming predictions, the theory whose prediction better aligns with human behavior better characterizes the representations humans build. In this work we introduce a new parser, the Serial Parser in ACT-R With Null elements (SPAWN), that can generate quantitative priming predictions from theories in syntax.

SPAWN is a cognitively motivated parser in which the parsing decisions are driven by the computational principles proposed by a general purpose cognitive architecture, Adaptive Control of Thought-Rational (ACT-R; \citealp{anderson2004}). Thus, 
SPAWN not only describes the computations underlying human parsing,
but also specifies the cognitive processes involved. 
This level of specification makes it possible to explain \textit{why}, given a grammar, some sentence A is primed more by sentence B compared to C, which in turn is useful for generating quantitative behavioral priming predictions from syntactic theories. 


Existing algorithmic models of parsing with this level of specification \cite{lewis2005} are limited in their ability to model assumptions from theories that use more contemporary frameworks like Minimalism for two reasons: First, they assume a disconnect between lexical and grammatical knowledge, which is inconsistent with the lexicalized grammar formalisms these frameworks adopt; Second, the models do not specify mechanisms to handle null (or covert) lexical items, which are essential components of several contemporary syntactic theories. SPAWN bridges this gap by adopting a lexicalized grammar formalism and specifying an explicit mechanism for null items.

As a case study, we use SPAWN to study the mental representations of sentences with relative clauses (RCs) such as \ref{ex:rrc1} and \ref{ex:frc1}. 
\vspace{-0.25em}
\ex. \label{ex:rrc1} The cat examined by the doctor was skittish. \vspace{-0.5em}

\ex. \label{ex:frc1} The cat which was examined by the doctor was skittish. \vspace{-0.5em}

We generate priming predictions from two competing syntactic theories: Whiz-Deletion \citep{chomsky1965}, which assumes that the structure of \ref{ex:rrc1} is identical to the structure of \ref{ex:frc1}, but that the words ``which'' and ``was'' are covert; and Participial-Phase \citep{harwood2018} which assumes that~\ref{ex:rrc1} and~\ref{ex:frc1} have different structures. We describe these theories in more detail in \S~\ref{sec:syntactic-theory}. We generate nine sets of predictions from the two theories by modulating two factors: First, the strength of prior knowledge (model exposed to 0, 100 or 1000 sentences before the experiment); Second, the reanalysis mechanism (model goes back to the beginning of the sentence, or model uses one of two entropy-based measures to select a word to go back to). Then, we compare the predictions from these two theories to empirical human data we collected using a novel web-based comprehension-to-production priming paradigm. 

We found that the predictions from the Whiz-Deletion never aligned with the qualitative pattern of human priming behavior, whereas under some assumptions about the underlying reanalysis mechanism and strength of prior knowledge, predictions from the Participial-Phase theory did align with the qualitative empirical pattern. These results suggest that the Participial-Phase account better characterizes human sentence representations. More broadly, this case study highlights how SPAWN can be used to adjudicate between competing theoretical assumptions: the quantitative behavioral predictions SPAWN generates can clarify how differences in assumptions about sentence structure or parsing mechanisms might translate into testable behavioral differences (if at all).



\begin{figure}
    \centering
    \includegraphics[width=\linewidth]{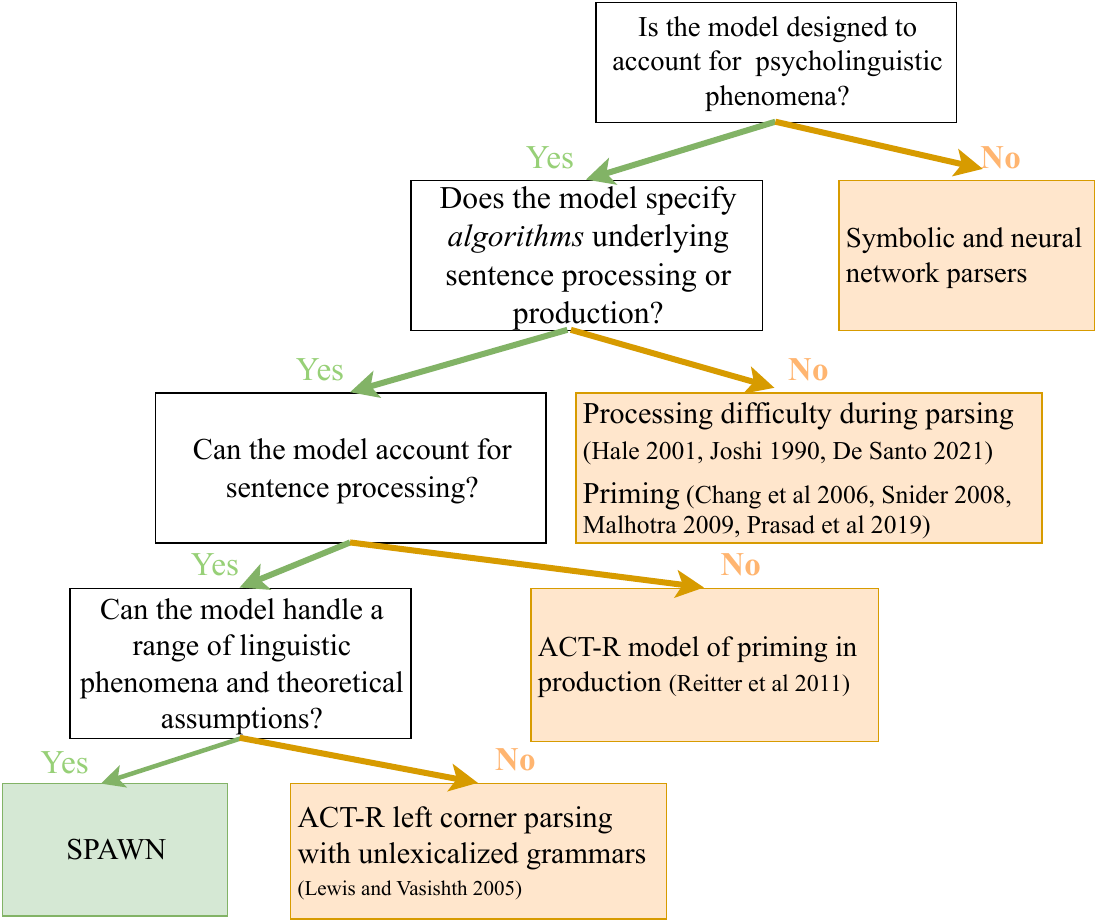}
    \caption{How is SPAWN different from other models?}
    \label{fig:why-spawn}
\end{figure}

\section{Background}
\subsection{The ACT-R framework}
ACT-R is a cognitive architecture designed to explain cognition through a small set of general computational principles and mechanisms that are relevant to a wide range of tasks and domains. One such mechanism which is particularly relevant in SPAWN is the retrieval of information from memory. The specific computational principles and algorithms that guide retrieval in ACT-R are outlined in \S~\ref{sec:retrieval}. Crucially, since ACT-R is intended to be a general purpose cognitive mechanism, most of the hyperparameters involved in this algorithm are already fixed based on data from a wide range of experimental paradigms and cognitive phenomena. This restricts the degrees of freedom and constrains the space of predictions that can be generated from any given theory.

\subsection{Prior models of parsing}
In most existing symbolic and neural-network based parsers, parsing decisions are not driven by specific cognitive principles such as the ones proposed by ACT-R. Therefore, generating predictions about observable human behavior (e.g., reading times) from these parsers requires making some additional \textit{linking hypotheses}. Most prior hypotheses that link parsing decisions to human behavior have focused on notions of processing effort, such as the number of parse states explored \cite{hale2011}, the maximum number of items on the stack at any given point \cite{joshi1990}, or the maximum amount of time a node stays in memory \cite{desanto2021minimalist}. These hypotheses cannot be used to generate priming predictions because they do not specify a mechanism by which a prime sentence might facilitate the processing of a target sentence.


One notable exception is the ACT-R based left-corner repair parser proposed by \citet{lewis2005}, in which parsing decisions are made based on the activation of different \textit{chunks} in the memory (such as words or grammar rules). The activation of chunks in this model can capture notions of both processing difficulty and priming. 
However, this model assumes a strong dissociation between the grammar and the lexicon and therefore cannot be adopted directly to generate predictions from lexicalized grammar formalisms such as Minimalist Grammar \cite{stabler1996derivational}, Combinatorial grammar \cite{steedman1988combinators}, Lexical-Functional Grammar \cite{kaplan1981lexical} or Head-Driven Phrase Structure Grammar \cite{pollard1994head}. SPAWN is an ACT-R parser that models the link between the grammar and the lexicon can therefore generate predictions from lexicalized grammars.

\subsection{Prior models of priming}
While several models of priming have been proposed, as we illustrate in Figure~\ref{fig:why-spawn}, none of them can be used to adjudicate between contemporary syntactic theories. Many models of priming that model sentence processing either do not explicitly model syntactic structure \citep{chang2006, malhotra09, prasad2019, sinclair2022structural} or do not explicitly implement the mechanisms that result in priming \citep{snider08}. \citet{reitter2011} proposed an ACT-R based model of priming that \textit{does} explicitly implement priming mechanisms and, unlike \citeauthor{lewis2005}'s ACT-R model, also assumes a strong link between lexical and grammatical knowledge, and is thus consistent with contemporary lexicalized grammar formalisms. However, this model can only generate sentences given a semantic description, and therefore can only be used to model sentence production and not sentence processing. We bridge this gap with SPAWN.


\section{Model description}
SPAWN uses the three components of ACT-R that are relevant for parsing: \textbf{declarative memory}, which contains information about lexical and syntactic categories (cf. \citealp{reitter2011}); \textbf{procedural memory}, which contains the algorithm for retrieving syntactic categories from memory and combining them together; and \textbf{buffers}, which store the words the parser has encountered so far, the syntactic categories retrieved for those words and the current parse state.\footnote{\url{https://github.com/grushaprasad/spawn}} We describe the two memory components below (\S~\ref{sec:declarative-memory}, \S~\ref{sec:procedural-memory}), as well as the mechanisms for learning and priming  (\S~\ref{sec:learning-in-spawn}, \S~\ref{sec:priming-in-spawn}).

\subsection{Declarative memory (\textit{the grammar})} \label{sec:declarative-memory}
Declarative memory in SPAWN consists of two types of \textit{chunks} (sets of attribute-value pairs): syntax chunks and lexical chunks (see \S~\ref{appendix:categories-shared} for the entire list of syntax and lexical chunks we use in this work).

\paragraph{Lexical chunks} Each lexical chunk stores a word in the vocabulary along with the set of syntactic categories that the word could be associated with. For example, 
the lexical chunk for ``examined'' encodes that it can be either be associated with the \textit{transitive verb} category or the \textit{past participle} category.

\paragraph{Syntax chunks} Each syntax chunk stores the constraints on the contexts in which a category can occur.
For example, 
the \textit{transitive verb} category encodes that it needs to have a \textit{determiner phrase} category on its left and right. We use the Combinatorial Categorial Grammar (CCG; \citealp{steedman1988combinators}) formalism to express such constraints.\footnote{The CCG notation to encode the ``transitive verb'' category is (TP$\backslash$DP)/DP; the forward slash indicates the words needs to combine with a DP on the right and the forward slash that it needs to combine with DP on the left. TP is the category that results from this combination.} 


\subsection{Procedural memory (\textit{the parser})}\label{sec:procedural-memory}
\begin{figure}
    \centering
    \includegraphics[width=\linewidth]{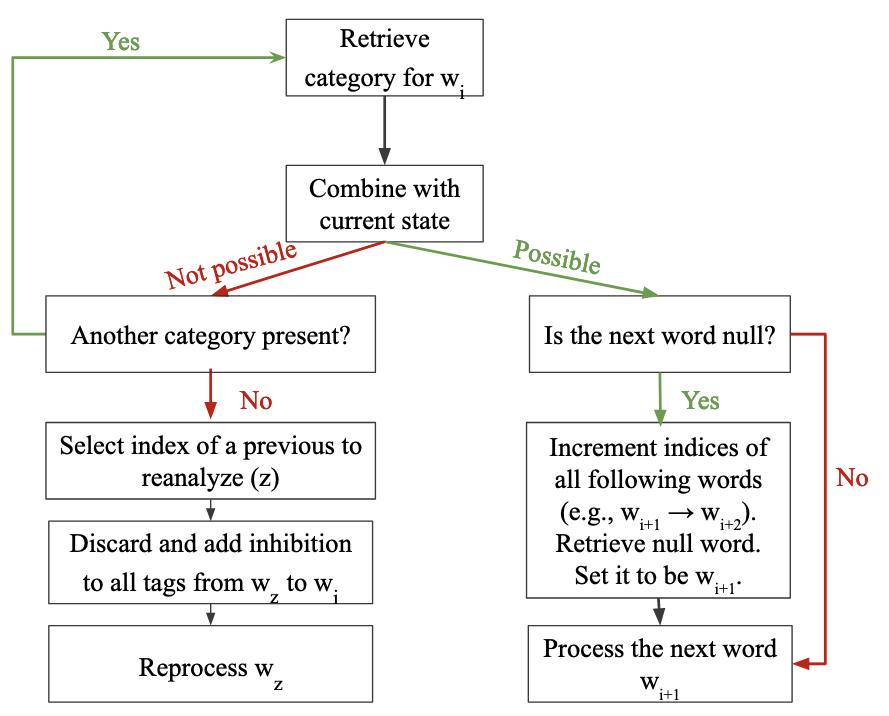}
    \caption{Steps involved in processing each word. Process is repeated till all words are assigned a category.} \vspace{-1em}
    \label{fig:procedural-memory-schematic}
\end{figure}
SPAWN parses sentences incrementally, one word at a time.  As schematized in Figure~\ref{fig:procedural-memory-schematic}, processing each word involves four steps: retrieval, reanalysis, integration and null-prediction. 



\subsubsection{Retrieval}\label{sec:retrieval}
When processing a word $w_i$ in a sentence $s$, the parser retrieves the category with the highest activation from the set $C_i$ of all possible categories that $w_i$ can be associated with. The activation $A_{ijs}$ for any category $c_{ij} \in C_i$ is given by Equation~\ref{eqn:activation}, where $B_{ij}$ is the base-level activation, $L_{ij}$ is the activation $w_i$ spreads to $c_{ij}$, $I_{ijs}$ is the inhibition from the buffer to the $c_{ij}$ when processing sentence $s$, and  $\epsilon$ is noise sampled from $\textit{Normal}(0, \sigma)$.
\vspace{-0.75em}
\begin{equation}\label{eqn:activation}
    A_{ijs} = B_{ij} + L_{ij} - I_{ijs} + \epsilon
\end{equation}
\vspace{-2em}

\paragraph{Base-level activation} This activation for a category is high if the category has been retrieved recently and/or frequently. It is given by Equation~\ref{eqn:base-act}, where $k$ is the total number of times the model has encountered $c_{ij}$, $T_{ijk}$ is the time taken to process all the words since the model’s $k$-th encounter of $c_{ij}$, and $d$ is a decay parameter. 
\vspace{-0.75em}
\begin{equation}\label{eqn:base-act}
    B_{ij} = \log \sum_{k=1}^K T_{ijk}^{-d} \vspace{-0.75em}
\end{equation}

The time to process a word $w_l$ is given by Equation~\ref{eqn:time-per-word}, where $N$ is the number of chunks retrieved when processing $w_{l}$, $A_{ln}$ the activation of the \mbox{$n$-th} chunk the model retrieved when processing $w_l$ (computed using Equation~\ref{eqn:activation}), $F$ a latency factor and $f$ a latency exponent. 

\vspace{-1em}
\begin{equation}\label{eqn:time-per-word}
     t_{l} = \sum_{n=1}^N F e^{-(f A_{ln})}
     \vspace{-0.5em}
\end{equation}

Thus, $T_{ij_k}$ in Equation~\ref{eqn:base-act} is $t_k +  t_{k+1} + \dots t_i$. 
   
    

\paragraph{Lexical activation}
The context independent activation a word $w_i$ spreads to a category $c_{ij}$ is given in Equation~\ref{eqn:lexical-act}, where $M$
is the maximum activation that any word can spread. 

\vspace{-0.75em}
\begin{equation} \label{eqn:lexical-act}
    L_{ij} = M\times P(c_{ij} \mid w_i) \vspace{-0.5em}
\end{equation}

\paragraph{Inhibition} The inhibition for $c_{ij}$ takes into account how often $c_{ij}$ was retrieved for $w_i$ but was later discarded during reanalysis when processing a sentence $s$. It increases if $c_{ij}$ was discarded often and/or recently, and is given by Equation~\ref{eqn:inhibition} where $Z$ indicates the total number of times $c_{ij}$ was discarded when processing $w_i$ in the current sentence, $T_{ijs_z}$ indicates the time since the $z$-th time $c_{ij}$ was discarded in sentence $s$, and $d$ is the decay factor.

\vspace{-0.75em}
\begin{equation} \label{eqn:inhibition}
    I_{ijs} = \log \sum_{z=1}^Z T_{ijs_z}^{-d} \vspace{-0.5em}
\end{equation}

The hyperparameters used in the equations above --- $d$, $F$, $f$, $M$ --- are set based on prior ACT-R models (\S~\ref{sec:priming-predictions}; \S~\ref{appendix:hyperparameters}). 

\subsubsection{Integration}\label{sec:integration}
Integrating a retrieved syntactic category $c_{ij}$ involves combining $c_{ij}$ with the current parse state $P$; this combination is determined by the CCG composition process (\citealt{steedman1996}; \S~\ref{appendix:ccg}). If no successful combination is possible, then the retrieved category cannot be integrated into the current parse state; the parser then needs to either retrieve another category for the word, or, if no unexplored categories remain, trigger a reanalysis.



\subsubsection{Reanalysis}\label{sec:reanalysis}
When a reanalysis gets triggered at $w_i$, the parser selects an index $z$ to regress to, where $z < i$. The method used to select $z$ is a hyperparameter with two settings: \textbf{first-word regression} (go back to the first word every time) and \textbf{entropy-weighted regression} (sample $z$ from $1 \dots i$ weighted by the parser's uncertainty at each index). Once the parser selects $z$, it discards all of the categories retrieved for $w_z \dots, w_{i-1}, w_{i}$, and resets the parse state to what it was at $w_{z}$. The parser keeps track of the categories that were discarded when processing each word in a sentence $s$, and uses this to compute the inhibition for each category using Equation~\ref{eqn:inhibition}.

\paragraph{Calculating uncertainty} To calculate uncertainty at index $x$ in entropy-weighted regression, we computed the activation of each category $c_{jx}$ associated with $w_x$ by adding together $B_{xj}$ and $L_{xj}$ (\S~\ref{sec:retrieval}). Then, we converted these activation values into probabilities with the softmax function (temperature 1 or 10), and finally computed the entropy from these probabilities.



\paragraph{``Give-up'' mechanism} \vspace{-0.4em}Despite inhibiting previously discarded categories, the parser could still get stuck in a loop retrieving the same (incorrect) category $c_{ij}$ every time it is processing $w_i$ if $c_{ij}$ has a very high base-level or lexical activation. To prevent an infinite loop, we implemented a ``give-up'' mechanism, where after $x$ iterations, the model ignores the base-level and lexical activation and uses only inhibition and noise to compute activation of $c_{ij}$. Setting $x$ to 100 or 1000 resulted in nearly identical results (\S~\ref{appendix:hyperparameters}).


\subsubsection{Null or covert element prediction}\label{sec:null-elements}
Null or covert elements in sentences
add additional uncertainty to the parsing process. To illustrate this, let us consider an example that is unrelated to our experimental setup, but illustrates the uncertainty in a theory-independent way. Given a prefix ``The cat examined the doctor and the doctor ...'' consider the following continuations; * indicates the continuation is ungrammatical. 




\vspace{-0.5em}
\ex. \label{ex:no-gap} ... \textbf{examined} the cat. \vspace{-0.25em}

\ex. \label{ex:gap}... \textcolor{red}{\textbf{NULL\textsubscript{examined}}} the cat. \vspace{-0.25em}

\ex. \label{ex:nogap-null} * ... \textcolor{red}{\textbf{NULL\textsubscript{examined}}} \textbf{examined} the cat. \vspace{-0.5em}

The covert \textcolor{red}{NULL\textsubscript{examined}} can only occur if its overt counterpart is absent. Therefore, after parsing the prefix, a serial parser has to predict whether the upcoming word in the sentence is covert or overt. If it expects the next word to be overt ``examined'', the parser should not retrieve any null elements. On the other hand, if it expects the next word to be covert \textcolor{red}{NULL\textsubscript{examined}}, it needs to retrieve this category and integrate it with the current parse state before processing the remainder of the sentence.  



We model this decision in SPAWN in the same way that we model other uncertainty: pick the option $N_{iks}$ with the highest activation, where $i$ is the current word, and $k \in \{x_1, x_2 \dots x_p, \textit{not-null}\}$, where $x_1, ..., x_p$ are the types of null elements that can come after the current parse state. The activation for $N_{iks}$ is given by Equation~\ref{eqn:null-act}:
\vspace{-0.5em}
\begin{equation}\label{eqn:null-act}
    N_{iks} = L_{ik} - I_{iks} + \epsilon \vspace{-0.5em}
\end{equation}

$L_{ik}$ and $I_{iks}$ are the same as in Equations~\ref{eqn:lexical-act} and~\ref{eqn:inhibition}. As in Equation~\ref{eqn:activation}, $\epsilon$ is noise sampled from $\textit{Normal}(0, \sigma)$. 
We do not include base-level activation for the null categories in this computation, because the base-level activation for the \textit{not-null} category would be extremely high (most sentences in the corpus do not have null elements) and would result in the null categories never being retrieved. We also assume that only certain parse states can be followed by null elements (\S~\ref{appendix:null-rules}): if the parser tried to insert null elements after every word, it would result in an exponential increase in the search space.

\subsection{Updating activations (``learning'')} \label{sec:learning-in-spawn}
Learning in SPAWN occurs by updating the counts of syntactic categories, which in turn are used to compute base-level and lexical activations (Equations~\ref{eqn:base-act}, \ref{eqn:lexical-act}). These counts are updated at the end of processing each sentence based on the final set of categories and null-elements that were retrieved. 

\subsection{Emergence of priming in SPAWN} \label{sec:priming-in-spawn}
Priming in SPAWN emerges as a consequence of parsing and learning. There are two factors that can result in priming: an increase in the activation of relevant categories and an increase in the probability of reanalysis. 

\paragraph{Increased activation} 
When a word in the target sentence is ambiguous between two categories $X$ and $Y$, if the parser retrieved $X$ in a preceding prime sentence, that increases its base and lexical activation relative to $Y$, which makes $X$ more likely to be retrieved in the target as well. 

\paragraph{Increased reanalysis} When a word in the target sentence is ambiguous between two categories $X$ and $Y$, and $Y$ has higher base and lexical activation, then the parser is more likely to retrieve $Y$ initially. If a sequence of parsing decisions causes the parser to reanalyze the word, then the probability of the parser eventually retrieving $X$ increases: the inhibition to $Y$ during reanalysis decreases the difference in activation between $X$ and $Y$.

\section{A case study: Evaluating competing theories of reduced relative clauses}\label{sec:syntactic-theory}
We use SPAWN to generate and test priming predictions for two competing syntactic theories of relative clauses that differ in their assumptions about how the structure of sentences like \ref{ex:rrc2} is related to the structure of sentences like \ref{ex:frc2} and \ref{ex:progrrc2}. 

\ex. \label{ex:rrc2} The cat examined by the doctor was skittish. (Reduced passive RC; RRC)

\ex. \label{ex:frc2} The cat who was examined by the doctor was skittish. (Full passive RC; FRC)

\ex. \label{ex:progrrc2} The cat being examined by the doctor was skittish. (Reduced progressive RC; ProgRRC)

Under the \textbf{Whiz-Deletion account} of RCs \citep{chomsky1965}, the sub-tree corresponding to any RC, whether reduced or not, is headed by the same node: a complementizer phrase (CP). In full RCs, the lexical content in this phrase (the wh-word and auxiliary ``was'') is overt, whereas in reduced RCs this lexical content is covert. By contrast, under the \textbf{Participial-Phase account} \citep{harwood2018}, while full RCs are headed by CPs, 
reduced passive and progressive RCs, are headed by Voice Phrase (VoiceP) and Progressive Phrase (ProgP) respectively. 
Consequently, the Participial-Phrase account, unlike the Whiz-Deletion account, does not assume the presence of a covert \textit{wh}-word and auxiliary in reduced passive and progressive RCs. See \S~\ref{appendix:trees} for trees that illustrate these differences.

\paragraph{Implementing the two theories}
We implement two versions of SPAWN, a Whiz-Deletion version and a Participial-Phase version. The procedural memory (parsing mechanism) is identical across both versions. There are two main differences in the declarative memory (grammar) across the versions. First, they differ in the categories that nouns can be associated with: in the Whiz-Deletion version, all nouns modified by RCs have the category \textit{NP/CP} (i.e., a noun looking to combine with a CP on its right), whereas in the Participial-Phase version, nouns modified by FRCs, RRCs and ProgRRCs are associated with different categories (\textit{NP/CP}, \textit{NP/VoiceP} and \textit{NP/ProgP} respectively). Second, the versions have different null lexical items: the Whiz-Deletion version has lexical items for a null subject Wh-word, a null finite auxiliary and a null progressive auxiliary, all of which are absent in the Participial-Phase version (see \S~\ref{appendix:categories-differences}).



\section{Methods}
\subsection{Experimental paradigm}
We used a comprehension-to-production priming paradigm to evaluate the two theories. In each experimental trial, human participants or SPAWN models were presented with three primes with the same structure, followed by an ambiguous partial prompt such as \ref{ex:target1} that could be completed either with or without a reduced RC. We used four prime types: three prime types with RCs (one each for RRC, FRC and ProgRRC), as well as control primes without RCs, such as  \ref{ex:amv-prime1}--\ref{ex:amv-prime3}.

\setlength{\Exlabelwidth}{0.5em}

\vspace{-0.5em}
\ex. \label{ex:amv-prime1} The dog chased the boy and ran away. \vspace{-0.25em}

\ex. \label{ex:amv-prime2} The monkey chased the hatter and stole a hat.\vspace{-0.25em}

\ex. \label{ex:amv-prime3} The dentist chased her son and panted. \vspace{-0.25em}

\ex. \label{ex:target1} The thief chased \_\_\_ 
\vspace{-0.5em}

\paragraph{Estimating priming effects} 
We estimated priming effects by measuring the proportion of RRC target parses in the different priming conditions (see \S~\ref{sec:human-exp} and \S~\ref{sec:priming-predictions} for details on how these parses were measured in humans and models). Concretely, we estimated $P(\textit{RRC parse} \mid \textit{target}, \textit{primes})$ by fitting Bayesian mixed-effects logistic regression model with the following three predictors (specified using Helmert contrasts) as fixed effects: All RCs vs. AMV, ProgRRC and FRC vs. RRC, and ProgRRC vs. FRC. We used a weakly informative prior and a maximal random effects structure (see \S~\ref{appendix:stats} for further details).

\paragraph{Materials}
When creating our stimuli, we picked 24 target verbs that can give rise to a temporary ambiguity as in \ref{ex:target1} which can either be resolved with either a main verb or reduced RC continuation. We created four items per verb and four versions of each item. The four versions of one of the items for the verb ``chased'' are illustrated below. 
\vspace{-0.5em}
\ex. \label{ex:chased-rrc} The dog chased by the boy ran away.\vspace{-0.25em}

\ex. \label{ex:chased-frc} The dog who was chased by the boy ran away.\vspace{-0.25em}

\ex. \label{ex:chased-progrrc} The dog being chased by the boy ran away.\vspace{-0.25em}

\ex. \label{ex:chased-amv} The dog chased the boy and ran away.
\vspace{-0.5em}

From these materials we created counterbalanced lists: in each list, three items occurred as primes; the fourth was cut at the verb to generate the target.

\subsection{Experiment with human participants}\label{sec:human-exp}
\paragraph{Participants}
We recruited 769 US-based participants from Prolific, of whom 765 were self-reported native speakers of English. We compensated them with 8.35 USD.  

\paragraph{Design}
We developed a web-based version of the comprehension-to-production priming paradigm used by \citet{pickering1998}. In the original paradigm, participants were given incomplete sentences in a booklet and asked to complete them. Since participants can be less attentive on web-based platforms than in the lab, we modified the paradigm to ensure that participants had to fully read the prime sentences. On the prime trials, participants were presented with a sentence, and asked to re-type that sentence from memory on the next screen. They could not progress until they typed in the sentence perfectly, and could not copy-paste the sentence, but could go back to re-read the sentence as often as they liked. On the target trials, participants were presented with the partial prompt on the screen, and asked to re-type the prompt and complete it on the next screen. They could not progress until they typed in the prompt perfectly and entered at least one more word. We did not automatically verify participants' productions, but in practice almost all participants generated grammatical completions with real words. 

\paragraph{Measuring the proportion of RRC parses}
We used regular expressions (\S~\ref{appendix:regex}) to classify all target completions into two categories (RRC vs. non-RRC) and specified RRC completions as ``success'' in our Bayesian logistic regression model. 

\begin{figure*}
    \centering
    \includegraphics[width=0.78\textwidth]{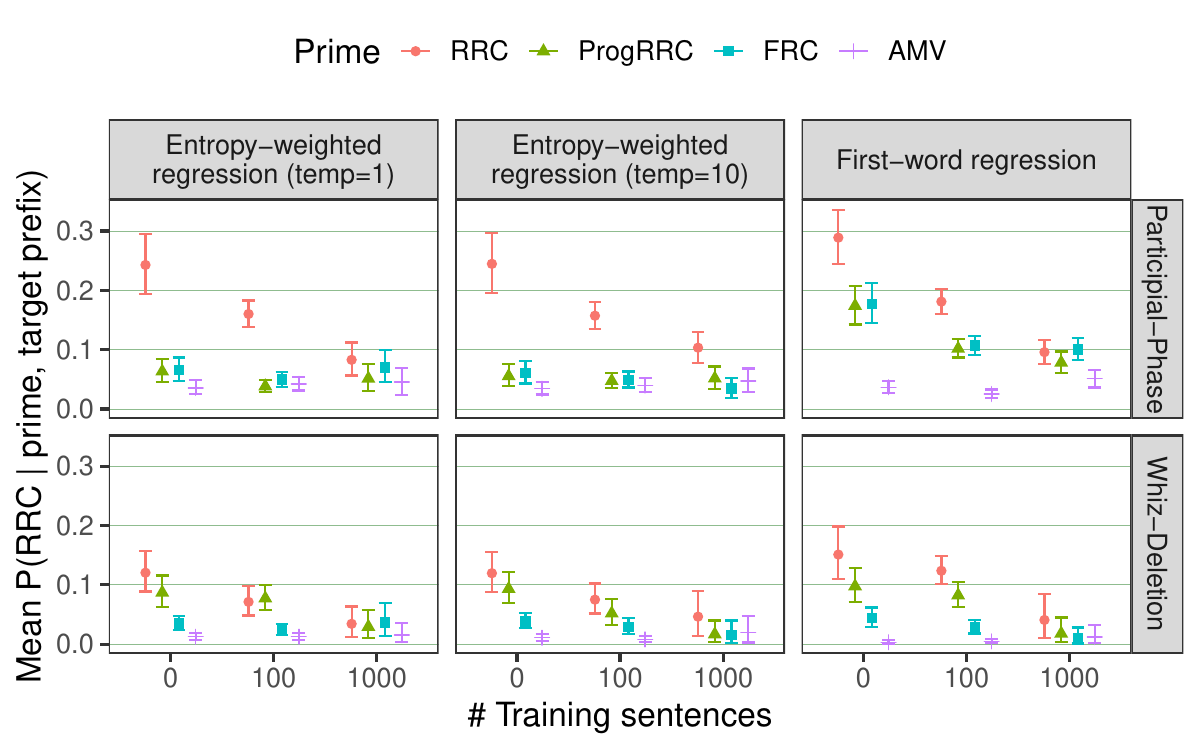}
    \caption{Predicted probability of RRC parse from the posterior distribution of the Bayesian logistic regression model. Error bars reflect 95\% credible intervals.}
    \label{fig:priming-preds}
\end{figure*}

\begin{figure}
    \centering
    \vspace{-1em}
    \includegraphics[width=0.75\linewidth]{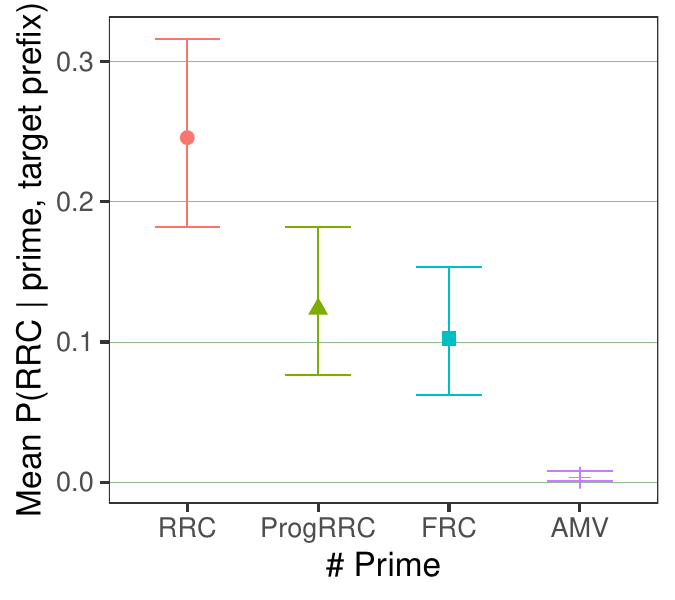}
    \caption{Empirical probability of RRC parse from the posterior distribution of the Bayesian logistic regression model. Error bars reflect 95\% credible intervals.} \vspace{-1.25em}
    \label{fig:priming-empirical}
\end{figure}

\subsection{Experiment with SPAWN models}\label{sec:priming-predictions}
We generated predictions from 18 types of models which varied along 3 dimensions: \textbf{the grammar} (Whiz-Deletion vs. Participial-Phase; \S~\ref{sec:syntactic-theory}, \ref{appendix:categories-differences}), \textbf{the reanalysis implementation} (First-word regression and  Entropy-Weighted reanalysis with temperature~1 or~10;  \S~\ref{sec:reanalysis}), and \textbf{the number of training sentences} (0, 100 or 1000 sentences).
For each model type, we created 1280 model instances, which, as we describe below, share some hyperparameters and differ in others.


\paragraph{Model hyperparameters}
The following hyperparameters are fixed across all model instances: 
decay ($d$  in Equations \ref{eqn:base-act},~\ref{eqn:inhibition}), latency exponent ($f$ in Equation \ref{eqn:time-per-word}), and maximum activation ($M$ in Equation \ref{eqn:lexical-act}). The following hyperparameters differ for each model instance: latency factor ($F$ in Equation~\ref{eqn:time-per-word}), and the noise parameter ($\sigma$ in \S~\ref{sec:retrieval}).

The values for $d$, $f$, and $M$ as well as the sampling distributions for $F$ and $\sigma$ were were taken from \citet{vasishth2021} (see \S~\ref{appendix:hyperparameters} for more details). We sampled $\sigma$ from $Normal(0.35, 1)$, because when sigma was sampled from \citeauthor{vasishth2021}'s $\textit{Uniform}(0.2, 0.5)$, some models never retrieved syntactic categories with low base-level activation. However, there were no qualitative differences in results between the two distributions (see \S~\ref{appendix:hyperparameters})


\paragraph{Training data}\label{sec:training-data}
To set initial base-level and lexical activations of the models prior to the experiment, we trained the models on 0, 100, or 1000 sentences and updated the activations at the end of each sentence as described in \S~\ref{sec:learning-in-spawn}.
These small numbers are consistent with prior work which assumes that participants start experiments with very weak priors \citep{delaney2019, fine2010}.

We used templates\footnote{\url{https://github.com/grushaprasad/spawn/blob/main/create_training_dat.py}} to generate a dataset of 10000 sentences in which the relative frequency of different types of RC sentences mirrored corpus statistics from \citet{roland2007}; for example, only 1\% of the training sentences contained an RRC (see \S~\ref{appendix:training} for details about the distribution of sentence types).
For each model instance, we sampled the training sentences from this dataset. Given the low probability of RRCs, many model instances never encountered these in their training data, and as such started with a base-level activation of $0$ for RRCs.

\paragraph{Design}
We presented each model instance with the stimuli from the human experiment. On a prime trial, the model parsed the sentence and updated the base-level and lexical activations based on the final set of retrieved categories. On a target trial, the model parsed the partial prompt and we recorded the resulting parse state; the model was constrained to end with only one of two partial states: DP/PP (RRC parse) or TP/DP (active parse).

\paragraph{Measuring the proportion of RRC parses}
We specified the DP/PP state as ``success'' in our Bayesian logistic regression model; unlike with humans, we do not need target completions to infer the parse the model assigned to the target.




\section{Results}


\paragraph{Participant/model exclusion} Most participants (77\%) never generated a single RRC target completion. Similarly most models (median of 67\% across the 18 model types) never generated a single RRC parse state. Since the goal of this work is to find \textit{differences} in the proportion of RRC parses between the primes, we only included in our analyses and plots the participants or models that generated at least one RRC completion or parse state. 


\paragraph{Human priming behavior}
In the human experiment, we observed that the proportion of target RRC parses was highest when the target was preceded by RRC primes with the same structure, and lowest when preceded by AMV primes which did not have any relative clauses. The proportion of target RRC parses in other two priming conditions, ProgRRC and FRC, were equivalent relative to each other, lower than with RRC primes, and higher than with AMV primes. (Figure~\ref{fig:priming-empirical}). See \S~\ref{appendix:statistical-inference} for statistical analyses.




\paragraph{Whiz-Deletion vs. Humans}
In the Whiz-Deletion models, processing ProgRRC sentences involves the retrieval of the same null complementizer as in RRC sentences, whereas processing FRC sentences does not (\S~\ref{appendix:sentences-analysis}). Consequently, these models predicted that the proportion of target RRC parses was greater with ProgRRC primes than with FRC primes (Figure~\ref{fig:priming-preds}), a pattern that does not align with the qualitative priming pattern observed in humans. Additionally, the magnitude of priming effects in the RRC condition were also generally smaller than what was observed in humans (Figure~\ref{fig:priming-preds}, bottom panel). These results together suggest that the Whiz-Deletion account of RRCs, at least the way we operationalized it, is not consistent with the representations humans build.

\paragraph{Participial-Phase vs. Humans}
In the Participial-Phase models, unlike in their Whiz-Deletion counterparts, processing ProgRRC, FRC, or AMV primes does not involve retrieving any categories that are shared with RRC sentences. However, the categories retrieved for ProgRRC and FRC but not AMV primes, increase the probability of reanalysis when processing the ambiguous target sentences (\S~\ref{appendix:sentences-analysis}). This reanalysis, as discussed in  \S~\ref{sec:priming-in-spawn}, in turn increases the probability of the model eventually assigning an RRC parse to the target, especially if the models' prior preference for AMV parses is relatively weak. Consequently, these models, particularly when they were trained on 0 and 100 sentences, predicted a graded effect which aligned with the qualitative priming pattern observed in humans: the proportion of target RRC parses was highest with RRC primes, followed by ProgRRC and FRC primes, and lowest with AMV primes. The models trained on $1000$ sentences could not capture this qualitative pattern because they generated very few RRC sentences across the board. This suggests, in line with prior work \citep{delaney2019, fine2010}, that when modeling the production or processing of extremely infrequent structures (like RRCs), assuming weak prior knowledge might be necessary.

Of the models that captured the qualitative patterns, the models with first-word regression better captured the \textit{magnitude} of the empirical priming effects (Figure~\ref{fig:priming-preds}).  Taken together, these results suggest that, depending on the assumptions we make about reanalysis and strength of prior belief, the Participial-Phase account of RRCs, unlike the Whiz-Deletion account, can be consistent with the representations humans build. 




\section{Discussion}
In this work we introduced a cognitively motivated parser, SPAWN, which can be used to generate quantitative behavioral predictions from contemporary syntactic theories that are based on lexicalized grammar formalisms. 
SPAWN makes it possible to evaluate what theoretical differences (if any) result in differing sentence processing predictions. 
As a case study, we used SPAWN to generate predictions from two competing theories of reduced relative clauses (Whiz-Deletion and Participial-Phase) while modulating the reanalysis mechanism and the number of training examples. We compared the predictions from these different versions of the SPAWN model to human behavior from a large-scale (N=769) web-based comprehension-to-production priming experiment. 

We found that the predictions of the Whiz-Deletion SPAWN models did not capture the qualitative human priming behavior for any of the model types. In contrast, many of the Participial-Phase SPAWN models captured the qualitative patterns, with the models that best captured the \textit{magnitude} of the empirical effects being ones with weak prior knowledge, that reprocesses the sentence from the beginning whenever reanalysis is triggered. Taken together, these results suggest that the Participial-Phase account of reduced relative clauses captures the structural representations people construct better than the Whiz-Deletion account.




\paragraph{Future work}
This work tentatively suggests that first-word regression might better model human processing than entropy-weighted regression. This observation needs to be more robustly validated with other empirical phenomena (e.g., priming in PO/DO sentences). Additionally, some of the parsing mechanisms SPAWN implements, such as for reanalysis or predicting null elements, are likely too simplistic to account for human sentence processing more generally (see \S~\ref{appendix:limitations}). Future work can tweak these mechanisms and evaluate the modified models against processing benchmarks like the SAP Benchmark \cite{huang2024large} which have more fine-grained measurements (e.g., reading time per word) across a range of psycholinguistic phenomena. Since the time taken for any of the parsing steps is measured in milliseconds by default in ACT-R, SPAWN can already generate quantitative predictions about the time taken to read or reprocess specific words in sentences, and therefore can be used with self-paced reading and eye-tracking datasets. Finally, future work can also use this paradigm to evaluate other competing syntactic theories. 


\paragraph{Conclusion} We proposed a cognitively plausible parser that can be used to generate \textit{quantitative} behavioral predictions from syntactic theories. Using English reduced relative clauses as a case study, we demonstrated how this model can be used to adjudicate between competing syntactic theories and parsing mechanisms. 

\section*{Acknowledgements}
We would like to thank the anonymous reviewers, HSP 2023 and 2024 audience as well as Aniello
De Santo, Shravan Vasishth, Will Merrill, Matt Wagers, Suhas Arehalli, Brian Dillon, Vijay Ramachandran and Joel Sommers for their valuable feedback. 

This work was partly supported by an American Psychological Association Dissertation Research Award. The work was conducted using computational resources from the Maryland Advanced Research Computing Center (MARCC) and the Colgate Supercomputer (Partially funded by NSF Award \#2346664). 
\bibliography{anthology,custom}
\bibliographystyle{acl_natbib}

\appendix
\onecolumn

\section{Details about the two theories of reduced relative clauses and how they are implemented in the declarative memory}
\subsection{Syntax trees}\label{appendix:trees}
\begin{figure}[h]
    \centering
    \begin{subfigure}[b]{0.6\textwidth}
        \centering
         \begin{tikzpicture}[scale=0.65]
        \Tree[.DP  [.\zero{D}\\the ] [.NP  [.\zero{N}\\cat ]  [.ProgP  [.\node(a){\zero{Prog}\\being}; ] [.vP  [.\node(b){\zero{v}\\$t_\textit{BE}$}; ] [.VoiceP  [.\node(x){\zero{Voice}}; ] [.VP  [.VP [.\node(y){\zero{V}\\examined}; ] ] [.\node(PP){PP}; \edge[roof]; {by the doctor}; ] ] ] ] ] ] ]      
        
        \draw [<-] (a) to [bend right = 50] (b);
        \draw[dashed] (y) to [bend left = 50] (x);
        \end{tikzpicture}
        \caption{Participial-Phase}
    \end{subfigure}
    \begin{subfigure}[b]{0.6\textwidth}
            \centering
    \begin{tikzpicture}[scale=0.65]
    \Tree[.DP  [.\zero{D}\\the ] [.NP  [.\zero{N}\\cat ] [.CP [.\node(DP){DP}; \edge[roof]; \color{red}{which} ] [.\Prime{C} [.\zero{C} ] [.TP [.\node(b){t$_\textit{DP}$}; ] [.\Prime{T} [.\node(t){\zero{T}\\\color{red}{was}}; ] [.vPProg  [.\node(s){\zero{v}prog\\$t_\textit{BE}$}; ]  [.ProgP  [.\node(p){\zero{Prog}\\being}; ] [.vP  [.\node(q){\zero{v}\\$t_\textit{BE}$}; ] [.VoiceP  [.\node(x){\zero{Voice}}; ] [.VP  [.VP [.\node(y){\zero{V}\\examined}; ] [.\node(c){t$_\textit{DP}$}; ] ] [.\node(PP){PP}; \edge[roof]; {by the doctor} ] ] ] ] ] ] ] ] ] ] ] ] 
    \node (a) [below=0.7cm of DP] {};
    \draw [<-] (p) to [bend right = 50] (q);
    \draw [dashed] (y) to [bend left = 50] (x);
    \draw [<-] (b) to [bend right = 100] (c);
    \draw [<-] (a) to [bend right = 50] (b);
    \draw [<-] (t) to [bend right = 50] (s);
\end{tikzpicture}
\caption{Whiz-Deletion}
    \end{subfigure}
   
    \caption{Syntax tree for ``The cat being examined by the doctor...''. The words in red are unvoiced in the Whiz-Deletion account. The tree for ``The cat examined by the doctor ...'' is nearly identical but without the ProgP. In Participial-Phase VoiceP is the sister of \textit{cat}; in Whiz-Deletion vP is the sister of \textit{was}. }
    \label{fig:rprc_prog-tree}
\end{figure}
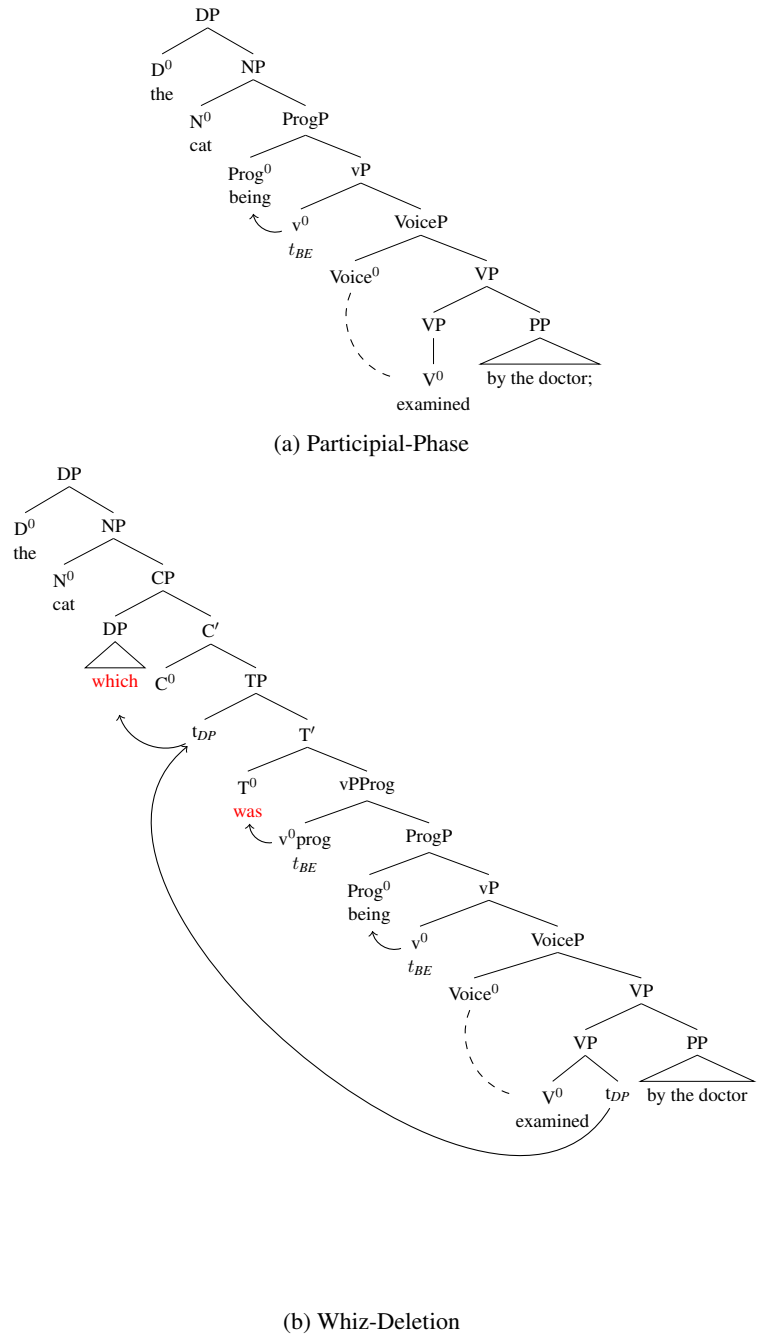
\newpage
\subsection{Differences in syntactic categories between the two theories}\label{appendix:categories-differences}
\begin{table}[h]
    \centering
    \resizebox{\textwidth}{!}{
    \begin{tabular}{llll}
    \toprule
       Category & Example sentence &  Whiz-Deletion 
         & Participial-Phase \\
         \midrule
         Noun & ``The cat which was examined by... '' & NP/CP & NP/CP \\
         (``cat'') & ``The cat examined by ... '' & NP/CP & NP/VoiceP \\
         & ``The cat being examined by ... '' & NP/CP & NP/ProgP\\
         \midrule
         Null wh subject & ``The cat NULL\textsubscript{wh} NULL\textsubscript{pass} examined by...'' &  CP/(TP$\backslash$DP) & MISSING \\
         Null finite auxiliary & & (TP$\backslash$DP)/VoiceP & MISSING \\
         Null progressive auxiliary & `The cat NULL\textsubscript{wh} NULL\textsubscript{prog} being examined by...'' & (TP$\backslash$DP)/ProgP & MISSING \\
         \bottomrule
    \end{tabular}
    }
    \label{tab:my_label}
\end{table}

\begin{figure*}
    \centering
    \begin{subfigure}[b]{0.95\textwidth}
        \centering
          \includegraphics[width=\textwidth]{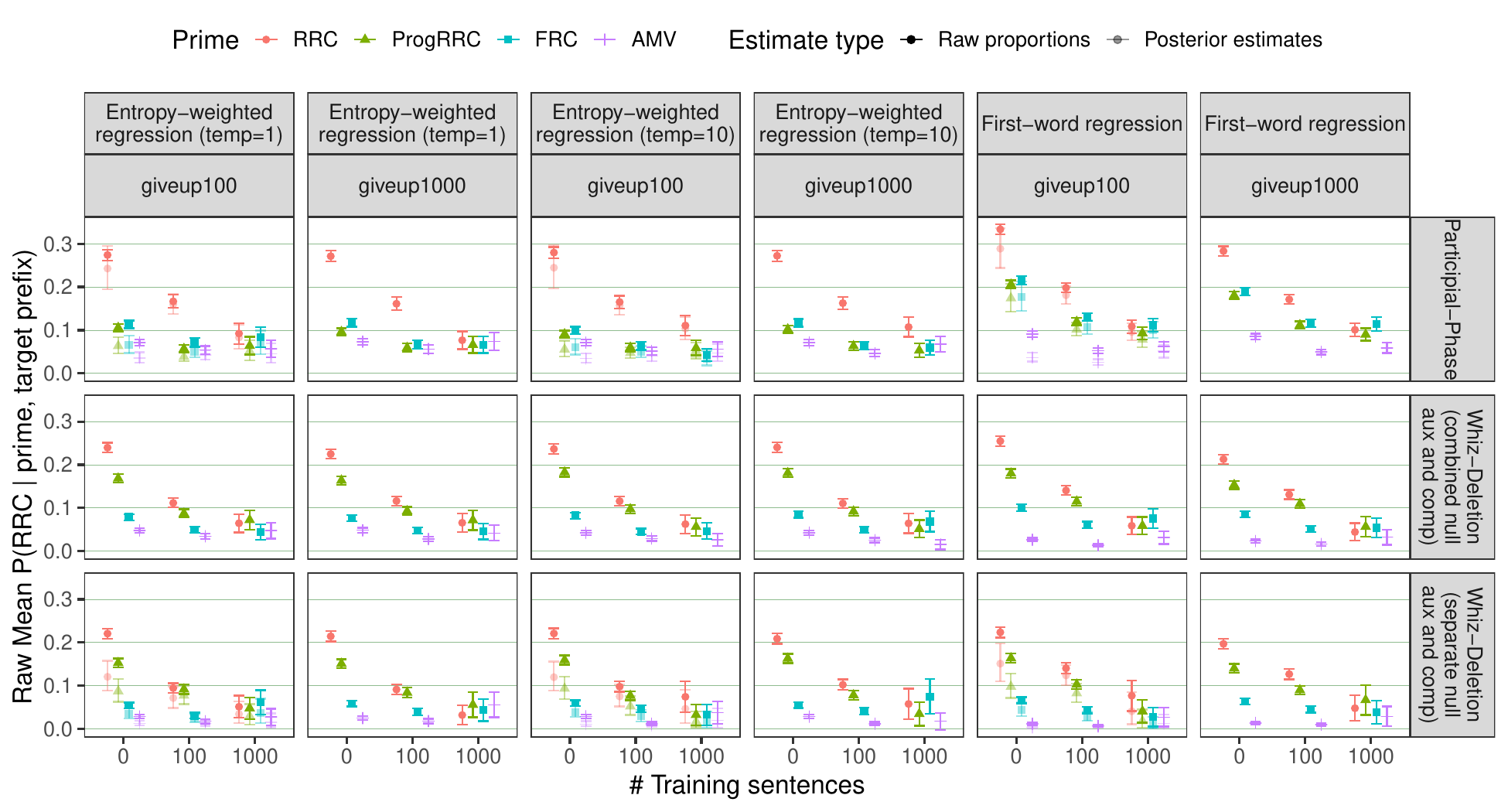}
         \caption{$\sigma$ sampled from $\textit{Normal}(0.35, 1)$}
         \label{fig:participial-phase-preds}
    \end{subfigure}
    \begin{subfigure}[b]{0.95\textwidth}
        \centering
          \includegraphics[width=\textwidth]{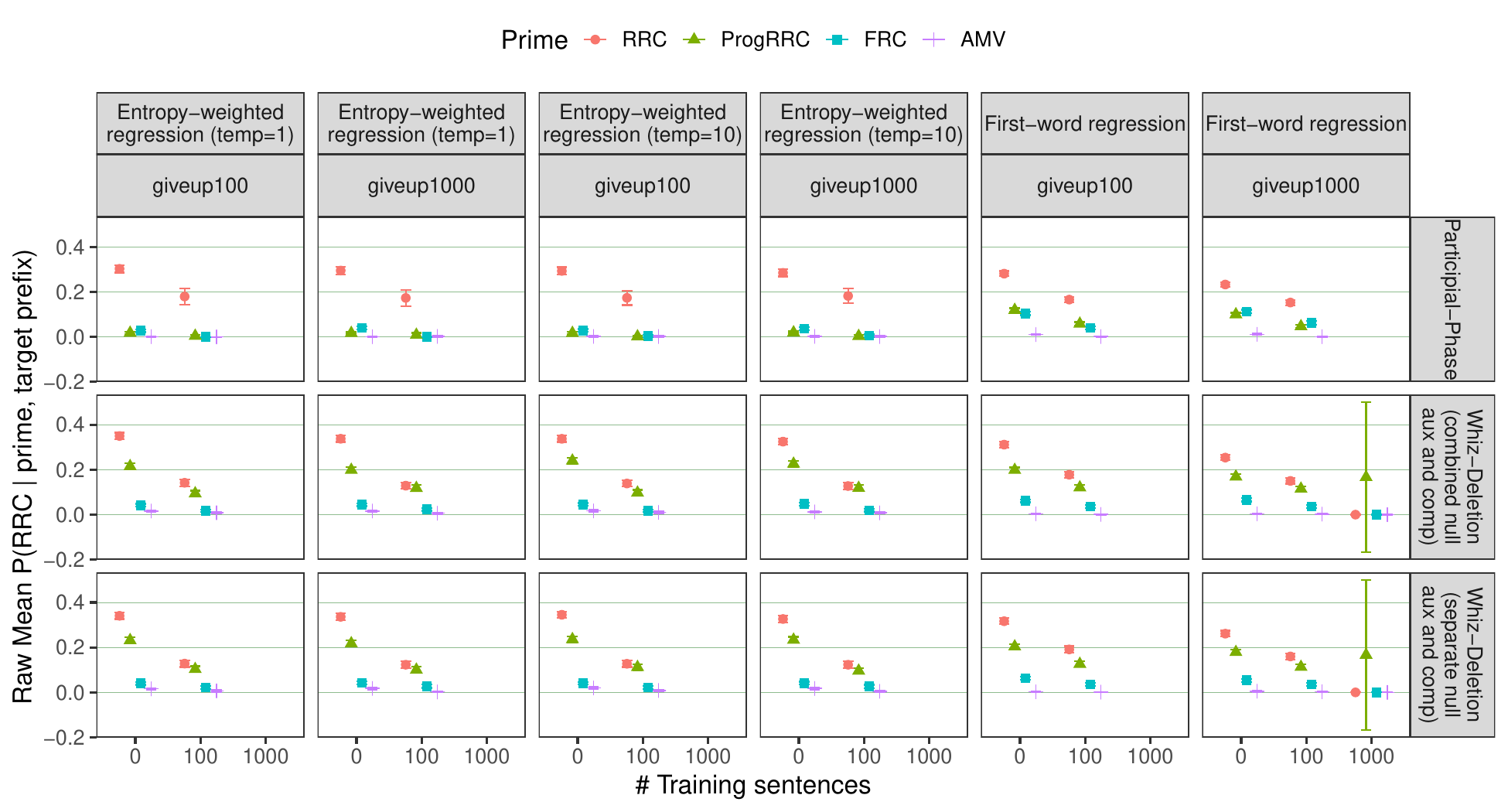}
         \caption{$\sigma$ sampled from $\textit{Uniform}(0.2, 0.5)$}
         \label{fig:whiz-del-preds}
    \end{subfigure}
    \caption{P(RRC $\mid$ prime, target) averaged across 1280 model instances as estimated with raw proportions (dark) and from the posterior distribution of Bayesian models (light). Since fitting Bayesian models is very time consuming, these models were fit only for the subset of results reported in the main text (Figure~\ref{fig:priming-preds}). Error bars represent 95\% standard error (standard deviation of proportions divided by $\sqrt{n}$) for proportions and 95\% Credible Intervals for the Bayesian models. Missing values indicate that no passive responses were generated. }\vspace{-1em}
    \label{fig:hyperparameters}
\end{figure*}

When the noun in the Whiz-Deletion version combines with the null Wh subject and null finite or progressive auxiliary, it results in the same parse state as the Participial-Phase noun categories for RRC and ProgRRC: NP/VoiceP and NP/ProgP. We also explored an alternative implementation of the Whiz-Deletion account where instead of having three NULL categories --- NULL\textsubscript{Wh}, NULL\textsubscript{pass} and NULL\textsubscript{prog} --- we had only two categories NULL\textsubscript{Whpass} (CP/VoiceP) and  NULL\textsubscript{Whprog} (CP/ProgP). This implementation resulted in nearly identical results (\S~\ref{fig:hyperparameters})

Note, both accounts have the same categories for the null wh-word in RCs which modify objects of clauses like ``The cat the doctor examined was skittish'': Null Complementizer (Object RCs) in the following table. 

\subsection{Syntactic categories shared by the two theories}\label{appendix:categories-shared}
\begin{table*}[h]
    \centering
    \resizebox{\textwidth}{!}{
    \begin{tabular}{lll}
    \toprule
    Category label     &  Example words & CCG rules \\
    \midrule
    Determiner & the, a , an, some, his, her, many, a-lot-of & DP/NP \\
    Determiner Phrase & something, everyone, non-violence, popularity & DP \\
    Noun Phrase & dragon, media, palace, mission, trance, tax-fraud & NP \\
    Preposition & on, to, into, by, at, in, down & PP/DP \\
    Transitive verb (active) &  accompanied, admired, betrayed, solved, forged & (TP$\backslash$DP)/DP \\
    Transitive verb (passive) & accompanied, admired, betrayed, solved, forged & VoiceP/PP \\
    Transitive verb (location object) &  arrived, staggered, marched, participated & (TP$\backslash$DP)/PP \\
    Intransitive verb & sang, cackled, complained, started-trending & TP$\backslash$DP \\
    Complementizer (Subject RC) & who & CP/(TP$\backslash$DP) \\
    Complementizer (Object RC) & who & CP/(((TP$\backslash$DP)/DP)/DP) \\
    Null Complementizer (Object RC) & NULL\textsubscript{Wh} & CP/(((TP$\backslash$DP)/DP)/DP) \\
    Prog & being & ProgP/VoiceP \\
    Auxiliary (followed by adjective) & was, were & (TP$\backslash$DP)/(NP/NP) \\
    Auxiliary (finite) & was & (TP$\backslash$DP)/VoiceP \\
     Auxiliary (progressive) & was & (TP$\backslash$DP)/ProgP \\
     Adjective & unreliable, competent, well-known, signature, radical & NP/NP \\
     Adverb & rapidly, diligently, in-surprise, sullenly, wistfully & TP$\backslash$TP \\
     Conjunction & and & (TP/(TP$\backslash$DP))$\backslash$TP \\
     EOS & . & end$\backslash$TP \\
     \bottomrule
    \end{tabular}
    }
    \caption{Categories present in the declarative memory in both Whiz-Deletion and Participial-Phase versions of SPAWN. In the syntax chunks, the category labels are the keys, and the CCG rules the attributes. In the lexical chunks, the words are the keys, and the category labels the attributes. The entire vocabulary can be found in the create\_declmem.py file in the Github repository.}
    \label{tab:shared-categories}
\end{table*}

\newpage
\section{SPAWN parsing details}


\subsection{CCG combination and type-raising rules}\label{appendix:ccg}
\begin{table}[h!]
    \centering
    \resizebox{0.8\textwidth}{!}{
    \begin{tabular}{lccc}
    \toprule
      \textbf{Rule name} & \textbf{Parser state form}  & \textbf{Tag form} & \textbf{Composed form}\\
      \midrule
      Forward composition & DP/NP  & NP & DP \\
      Backward composition & DP & TP$\backslash$DP & TP \\
      Forward harmonic composition & DP/VoiceP & VoiceP/PP & DP/PP \\
      Backward harmonic composition & TP$\backslash$DP & eos$\backslash$TP & eos$\backslash$DP \\
      Forward crossed composition & CP/TP & TP$\backslash$DP & CP$\backslash$DP \\
      Backward crossed composition & TP/VoiceP & eos$\backslash$TP & eos/VoiceP\\
      \bottomrule
    \end{tabular}
    }
    \caption{Examples of all the six possible CCG composition rules being applied when parsing sentences in the training set.}
    \label{tab:ccg_rules}
\end{table}

\noindent We have just one type-raising rule: DP can get type-raised to TP/(TP$\backslash$DP). This lets the subject DP in a sentence combine with a transitive verb --- (TP$\backslash$DP)/DP --- before the transitive verb combines with the object DP.

The parser starts by sequentially trying to apply each of the six composition rules, stopping once a successful combination is found. If no successful combination is found, then the parser tries to the type-raising rule and then sequentially apply all six composition rules. 

\subsection{Categories that can be followed by null elements}\label{appendix:null-rules}
In the Whiz-Deletion grammar the NP/CP category, and the CP/(TP$\backslash$DP) can be followed by null elements, whereas in the Participial-Phase grammar, only the the NP/CP category can be followed by a null element (to account for object reduced RCs like ``The cat the doctor examined was skittish''). 


\subsection{Analysis of example sentences with our grammar}\label{appendix:sentences-analysis}
\begin{table}[h]
    \centering
     \resizebox{0.9\textwidth}{!}{
    \begin{tabular}{lllll}
    \toprule
        \textbf{Old Parse state} & \textbf{Word} & \textbf{Correct category} & \textbf{Rule} & \textbf{New parse state}\\
        \midrule
        NULL & the & DP/NP & Initialize & DP/NP \\ 
        DP/NP & cat & NP/VoiceP & Forward Harmonic Composition & DP/VoiceP \\
        DP/VoiceP & examined & VoiceP/PP & Forward Harmonic Composition & DP/PP \\
        DP/PP & by & PP/DP & Forward Harmonic Composition & DP/DP \\
        DP/DP & the & DP/NP & Forward Harmonic Composition & DP/NP \\
        DP/NP & doctor & NP & Forward Composition & DP \\
        \color{gray}{DP} & \color{gray}{liked} & \color{gray}{(TP$\backslash$DP)/DP} & \color{gray}{Type raise DP}   \\ 
       \color{gray}{TP/(TP$\backslash$DP)} & \color{gray}{}& \color{gray}{} & \color{gray}{Forward Harmonic composition} & \color{gray}{TP/DP} \\
        \color{gray}{TP/DP} & \color{gray}{the} & \color{gray}{DP/NP} & \color{gray}{Forward Harmonic composition} & \color{gray}{TP/NP} \\
        \color{gray}{TP/NP} & \color{gray}{girl} & \color{gray}{NP} & \color{gray}{Forward Harmonic composition} & \color{gray}{TP} \\
        \color{gray}{TP} & \color{gray}{EOS} & \color{gray}{end$\backslash$TP} & \color{gray}{Backward composition} & \color{gray}{end}\\
        \bottomrule
    \end{tabular}
    }
    \caption{CCG analysis for a reduced RC sentence under the Participial-Phase grammar. The rows in gray are the same across all RC types. }
    \label{tab:ccg-analysis-PP-RRC}
\end{table}

\begin{table}[h!]
    \centering
     \resizebox{0.9\textwidth}{!}{
    \begin{tabular}{lllll}
    \toprule
        \textbf{Old Parse state} & \textbf{Word} & \textbf{Correct category} & \textbf{Rule} & \textbf{New parse state}\\
        \midrule
        NULL & the & DP/NP & Initialize & DP/NP \\ 
        DP/NP & cat & NP/CP & Forward Harmonic Composition & DP/CP \\
        DP/CP & \color{red}{NULL\textsubscript{wh}} & CP/(TP$\backslash$DP) & Forward Harmonic Composition & DP/(TP$\backslash$DP) \\
        DP/(TP$\backslash$DP) & \color{red}{NULL\textsubscript{pass}} & (TP$\backslash$DP)/VoiceP & Forward Harmonic Composition & DP/VoiceP \\
        DP/VoiceP & examined & VoiceP/PP & Forward Harmonic Composition & DP/PP \\
        DP/PP & by & PP/DP & Forward Harmonic Composition & DP/DP \\
        DP/DP & the & DP/NP & Forward Harmonic Composition & DP/NP \\
        DP/NP & doctor & NP & Forward Composition & DP \\
        \color{gray}{DP} & \color{gray}{liked} & \color{gray}{(TP$\backslash$DP)/DP} & \color{gray}{Type raise DP}   \\ 
       \color{gray}{TP/(TP$\backslash$DP)} & \color{gray}{}& \color{gray}{} & \color{gray}{Forward Harmonic composition} & \color{gray}{TP/DP} \\
        \color{gray}{TP/DP} & \color{gray}{the} & \color{gray}{DP/NP} & \color{gray}{Forward Harmonic composition} & \color{gray}{TP/NP} \\
        \color{gray}{TP/NP} & \color{gray}{girl} & \color{gray}{NP} & \color{gray}{Forward Harmonic composition} & \color{gray}{TP} \\
        \color{gray}{TP} & \color{gray}{EOS} & \color{gray}{end$\backslash$TP} & \color{gray}{Backward composition} & \color{gray}{end}\\
        \bottomrule
    \end{tabular}
    }
    \caption{CCG analysis for a reduced RC sentence under the Whiz-Deletion grammar. The rows in gray are the same across all RC types. We experimented with an alternative version where NULL\textsubscript{Wh} and NULL\textsubscript{pass} were combined into one category. This resulted in qualitatively similar results.}
    \label{tab:ccg-analysis-PP-RRC}
\end{table}

\begin{table}[h!]
    \centering
     \resizebox{0.9\textwidth}{!}{
    \begin{tabular}{lllll}
    \toprule
        \textbf{Old Parse state} & \textbf{Word} & \textbf{Correct category} & \textbf{Rule} & \textbf{New parse state}\\
        \midrule
        NULL & the & DP/NP & Initialize & DP/NP \\ 
        DP/NP & cat & NP/ProgP & Forward Harmonic Composition & DP/ProgP \\
        DP/ProgP & being & ProgP/VoiceP & Forward Harmonic Composition & DP/VoiceP \\
        DP/VoiceP & examined & VoiceP/PP & Forward Harmonic Composition & DP/PP \\
        DP/PP & by & PP/DP & Forward Harmonic Composition & DP/DP \\
        DP/DP & the & DP/NP & Forward Harmonic Composition & DP/NP \\
        DP/NP & doctor & NP & Forward Composition & DP \\
        \color{gray}{DP} & \color{gray}{liked} & \color{gray}{(TP$\backslash$DP)/DP} & \color{gray}{Type raise DP}   \\ 
       \color{gray}{TP/(TP$\backslash$DP)} & \color{gray}{}& \color{gray}{} & \color{gray}{Forward Harmonic composition} & \color{gray}{TP/DP} \\
        \color{gray}{TP/DP} & \color{gray}{the} & \color{gray}{DP/NP} & \color{gray}{Forward Harmonic composition} & \color{gray}{TP/NP} \\
        \color{gray}{TP/NP} & \color{gray}{girl} & \color{gray}{NP} & \color{gray}{Forward Harmonic composition} & \color{gray}{TP} \\
        \color{gray}{TP} & \color{gray}{EOS} & \color{gray}{end$\backslash$TP} & \color{gray}{Backward composition} & \color{gray}{end}\\
        \bottomrule
    \end{tabular}
    }
    \caption{CCG analysis for a reduced progressive RC sentence under the Participial-Phase grammar. The rows in gray are the same across all RC types. }
    \label{tab:ccg-analysis-PP-RRC}
\end{table}

\begin{table}[h!]
    \centering
     \resizebox{0.9\textwidth}{!}{
    \begin{tabular}{lllll}
    \toprule
        \textbf{Old Parse state} & \textbf{Word} & \textbf{Correct category} & \textbf{Rule} & \textbf{New parse state}\\
        \midrule
        NULL & the & DP/NP & Initialize & DP/NP \\ 
        DP/NP & cat & NP/CP & Forward Harmonic Composition & DP/CP \\
        DP/CP & \color{red}{NULL\textsubscript{wh}} & CP/(TP$\backslash$DP) & Forward Harmonic Composition & DP/(TP$\backslash$DP) \\
        DP/(TP$\backslash$DP) & \color{red}{NULL\textsubscript{prog}} & (TP$\backslash$DP)/ProgP & Forward Harmonic Composition & DP/VoiceP \\
        DP/ProgP & being & ProgP/VoiceP & Forward Harmonic Composition & DP/VoiceP \\
        DP/VoiceP & examined & VoiceP/PP & Forward Harmonic Composition & DP/PP \\
        DP/PP & by & PP/DP & Forward Harmonic Composition & DP/DP \\
        DP/DP & the & DP/NP & Forward Harmonic Composition & DP/NP \\
        DP/NP & doctor & NP & Forward Composition & DP \\
        \color{gray}{DP} & \color{gray}{liked} & \color{gray}{(TP$\backslash$DP)/DP} & \color{gray}{Type raise DP}   \\ 
       \color{gray}{TP/(TP$\backslash$DP)} & \color{gray}{}& \color{gray}{} & \color{gray}{Forward Harmonic composition} & \color{gray}{TP/DP} \\
        \color{gray}{TP/DP} & \color{gray}{the} & \color{gray}{DP/NP} & \color{gray}{Forward Harmonic composition} & \color{gray}{TP/NP} \\
        \color{gray}{TP/NP} & \color{gray}{girl} & \color{gray}{NP} & \color{gray}{Forward Harmonic composition} & \color{gray}{TP} \\
        \color{gray}{TP} & \color{gray}{EOS} & \color{gray}{end$\backslash$TP} & \color{gray}{Backward composition} & \color{gray}{end}\\
        \bottomrule
    \end{tabular}
    }
    \caption{CCG analysis for a reduced RC sentence under the Whiz-Deletion grammar. The rows in gray are the same across all RC types. We experimented with an alternative version where NULL\textsubscript{Wh} and NULL\textsubscript{pass} were combined into one category. This resulted in qualitatively similar results.}
    \label{tab:ccg-analysis-PP-RRC}
\end{table}

\begin{table}[h!]
    \centering
     \resizebox{0.9\textwidth}{!}{
    \begin{tabular}{lllll}
    \toprule
        \textbf{Old Parse state} & \textbf{Word} & \textbf{Correct category} & \textbf{Rule} & \textbf{New parse state}\\
        \midrule
        NULL & the & DP/NP & Initialize & DP/NP \\ 
        DP/NP & cat & NP/CP & Forward Harmonic Composition & DP/CP \\
        DP/CP & which & CP/(TP$\backslash$DP) & Forward Harmonic Composition & DP/(TP$\backslash$DP) \\
        DP/(TP$\backslash$DP) & was & (TP$\backslash$DP)/VoiceP & Forward Harmonic Composition & DP/VoiceP \\
        DP/VoiceP & examined & VoiceP/PP & Forward Harmonic Composition & DP/PP \\
        DP/PP & by & PP/DP & Forward Harmonic Composition & DP/DP \\
        DP/DP & the & DP/NP & Forward Harmonic Composition & DP/NP \\
        DP/NP & doctor & NP & Forward Composition & DP \\
        \color{gray}{DP} & \color{gray}{liked} & \color{gray}{(TP$\backslash$DP)/DP} & \color{gray}{Type raise DP}   \\ 
       \color{gray}{TP/(TP$\backslash$DP)} & \color{gray}{}& \color{gray}{} & \color{gray}{Forward Harmonic composition} & \color{gray}{TP/DP} \\
        \color{gray}{TP/DP} & \color{gray}{the} & \color{gray}{DP/NP} & \color{gray}{Forward Harmonic composition} & \color{gray}{TP/NP} \\
        \color{gray}{TP/NP} & \color{gray}{girl} & \color{gray}{NP} & \color{gray}{Forward Harmonic composition} & \color{gray}{TP} \\
        \color{gray}{TP} & \color{gray}{EOS} & \color{gray}{end$\backslash$TP} & \color{gray}{Backward composition} & \color{gray}{end}\\
        \bottomrule
    \end{tabular}
    }
    \caption{CCG analysis for a full passive RC sentence under the Whiz-Deletion and Participial-Phase grammar.The rows in gray are the same across all RC types.}
    \label{tab:ccg-analysis-PP-RRC}
\end{table}

\begin{table}[h!]
    \centering
     \resizebox{0.9\textwidth}{!}{
    \begin{tabular}{lllll}
    \toprule
        \textbf{Old Parse state} & \textbf{Word} & \textbf{Correct category} & \textbf{Rule} & \textbf{New parse state}\\
        \midrule
        NULL & the & DP/NP & Initialize & DP/NP \\ 
        DP/NP & cat & NP & Forward Composition & DP \\
         DP & examined & (TP$\backslash$DP)/DP & Type raise DP & \\
         TP/(TP$\backslash$DP) & & &Forward Harmonic Composition & TP/DP \\
         TP/DP & the & DP/NP & Forward Harmonic Composition & TP/NP \\
         TP/NP & doctor & NP & Forward Composition & TP \\
          TP & and & (TP/(TP$\backslash$DP))$\backslash$TP) & Backward composition & TP/(TP$\backslash$DP) \\
        (TP/(TP$\backslash$DP))$\backslash$TP) & liked & (TP$\backslash$DP)/DP & Forward Harmonic composition & TP/DP  \\ 
        \color{gray}{TP/DP} & \color{gray}{the} & \color{gray}{DP/NP} & \color{gray}{Forward Harmonic composition} & \color{gray}{TP/NP} \\
       TP/NP & girl & NP & Forward Harmonic composition & TP \\
        TP & EOS & end$\backslash$TP & Backward composition & end\\
        \bottomrule
    \end{tabular}
    }
    \caption{CCG analysis for an active main verb sentence with verb coordination under the Whiz-Deletion and Participial-Phase grammar.}
    \label{tab:ccg-analysis-PP-RRC}
\end{table}


\newpage
\section{Details about the training dataset}\label{appendix:training}
\begin{table}[h]
    \centering
    \resizebox{\textwidth}{!}{
    \begin{tabular}{lll}
    \toprule
       \textbf{Structure}  & \textbf{Prob}  &\textbf{Example}\\
      \midrule
       Subject RC  & 0.016 & The defendant who examined the lawyer ... \\
       Full object RC & 0.002 & The defendant who the lawyer examined ... \\
       Reduced object RC & 0.005 & The defendant the lawyer examined ... \\
       Full passive RC & 0.002 & The defendant who was examined by the lawyer ... \\
       Reduced passive RC & 0.011 & The defendant examined by the lawyer ... \\
       Full progressive RC & 0.0002 & The defendant who was being examined by the lawyer ... \\
       Reduced progressive RC & 0.005 & The defendant being examined by the lawyer ... \\
       \midrule
       Transitive NP object & 0.321 & The examined the lawyer. \\
       Transitive PP object & 0.080 & The defendant went to the store. \\
       Intransitive & 0.240 & The defendant sang (joyfully). \\
       Copular & 0.240 & The defendant was happy.\\
       \midrule
       Coordination & 0.080 & The defendant examined the lawyer and went to the store. 
       \\
       & & The defendant was happy and sang joyfully.\\
       & & The defendant went to the store and sang and was happy and examined the lawyer. \\
       
       \bottomrule
       
    \end{tabular}}
    \caption{The relative frequencies for all RCs, except the Progressive RCs, was taken from \cite{roland2007}. Since progressive RCs were absent from this corpus study, we approximated their probabilities informally using google n-grams: for a range of different verbs, full RCs with almost never showed up in google n-gram viewer, but progressive RCs occasionally did. So we set the probability of progressive RCs to be twice that of full RCs. Since reduced RCs were much more frequent than their full counterparts, we assigned 95\% of the probability mass of progressive RCs to the reduced version, and the remaining five to the full version. Since the exact frequencies of non-RC sentences is unlikely to be relevant for our experimental set up, we just included a few types of non-RC sentences without trying to match their frequencies with corpus statistics.}
\end{table}

\newpage
\section{Exploring model hyperpameters}\label{appendix:hyperparameters}
\begin{table*}[h!]
    \centering
    \resizebox{\textwidth}{!}{
    \begin{tabular}{llll}
    \toprule
        Hyperparameter & Equation or Section & Value(s) & Reason  \\
        \midrule
        Decay ($d$) & Eqn~\ref{eqn:base-act}, Eqn~\ref{eqn:inhibition} & 0.5 & \citet{vasishth2021} \\
        Latency exponent ($f$) & Eqn~\ref{eqn:time-per-word} & 1 & \citet{vasishth2021} \\
        Maximum actiation ($M$) & 1.5 & Eqn~\ref{eqn:lexical-act} &\citet{vasishth2021} \\
        \midrule
        Latency factor ($F$) & Eqn~\ref{eqn:time-per-word} & $\textit{Beta}(2,6)$ & \citet{vasishth2021} \\
        SD of noise distribution ($\sigma$) & \S~\ref{sec:retrieval}& $\textit{Uniform}(0.2, 0.5)$ & \citet{vasishth2021}\\
        & & $\textit{Normal}(0.35, 1)$ & Add more noise to retrieve passive. \\
        \midrule
        \midrule
        \# Training sentences & \S~\ref{sec:training-data} &  0, 100, 1000 & >1000 resulted in almost no passive retrieval. \\
        Give up & \S~\ref{sec:reanalysis} & 100, 1000 & >1000 too much time; 100,1000 same behavior. \\
        Reanalysis index ($z$) & \S~\ref{sec:reanalysis} & 1  & Always go back to first word. \\
        & & Entropy weighted sample; SM temp: 1 & Emphasize differences in activation. \\
        & & Entropy weighted sample; SM temp: 10 & Make differences in activation more uniform. \\
        \midrule
        Random seed ($s$) & & Between 1 to 1280 & Affects training order, random sampling. \\
        \bottomrule
    \end{tabular}
    }
    \caption{Hyperparameters above the double line are ACT-R parameters. Hyperparameters below the double line are SPAWN specific hyperparameters. Only $F$, $\sigma$ and $s$ differ across the 1280 model instances.}
    \label{tab:my_label}
\end{table*}

\newpage
\section{Details about statistical models}\label{appendix:stats}
\subsection{Model specification}
To generate quantitative predictions about the predicted proportion of passive responses while taking into consideration the model-instance wise and item wise variation, we fit Bayesian mixed effects logistic regression models. We used a Helmert contrast coding scheme with the following predictors, which let us evaluate if the mean log odds ratio of the ProgRRC and FRC conditions are equal to each other and to the mean log odds ratio of the RRC condition.

\begin{itemize}
    \item C1: Compare the mean log odds ratio of the AMV condition with the mean log odds ratio of all the RC conditions combined. 
    \item C2: Compare the mean log odds ratio of the RRC condition with the mean log odds ratio of all ProgRRC and FRC conditions combined.
    \item C3: Compare the mean log odds ratio of the ProgRRC condition to the mean log odds ratio of the FRC condition. 
\end{itemize}

We fit the maximal model by including all by-participant and by-item random intercepts and slopes. In the case of the predicted data, participant IDs were replaced by model instance IDs.  
\begin{align*}\label{eqn:brms-notation}
    Passive \sim \ &c1 + c2 + c3 + \\\vspace{-1em}
    &(1 + c1 + c2 + c3 \mid item) +\\\vspace{-1em}
    &(1 + c1 + c2 + c3 \mid \textit{participant or model-instance})
\end{align*}

\subsection{Priors}
We fit the models using the following weakly informative prior. 

\begin{align*}
    \textit{Intercept} &\sim Normal(-4.595, 1.5) \\\vspace{-1em}
    \textit{Fixed effects} &\sim Normal(0,2) \\\vspace{-2em}
   \textit{ SD for random effects} &\sim Normal(0,5)
\end{align*}

This prior assumes that the log odds ratio between priming conditions is most likely to be 0 (i.e. no priming effect) and unlikely to be greater than 4 or less than -4. This assumption is based on a meta-analysis of priming in production studies \citep{mahowald2016meta} where the log odds ratio between the prime conditions was not greater than 4 in any of the constructions they considered.

\subsection{Statistical inferences for empirical human data}\label{appendix:statistical-inference}
As discussed in the main text, we observed the following qualitative pattern in the proportion of target RRC parses when preceded by different primes: RRC >  ProgRRC = FRC > AMV. To ensure that this pattern was statistically valid, we computed Bayes Factors for all of our predictors using the bayestestR package \cite{bayestestR}. We adopt the Bayes Factor scale from \citet{jeffreys1998theory} to draw inferences: values greater than 3 and 10 provide moderate and strong evidence for the alternative model, whereas values lower than 0.3 and 0.1 provide moderate and strong evidence for the null model. Therefore, the following Bayes Factor values for our predictors would support the qualitative pattern: 
\begin{enumerate}
    \item AMV vs. all RCs (C1): > 3
    \item RRC vs. [ProgRRC and FRC] (C2): > 3
    \item ProgRRC vs. FRC (C3): < 0.3
\end{enumerate}

\begin{table}[h]
    \centering
    \begin{tabular}{llll}
    \toprule
         \textbf{Predictor} & \textbf{Estimate} & \textbf{95\% CI} &\textbf{ Bayes Factor}  \\
         \midrule 
         AMV vs. all RCs (C1) & -4.18 & [-5.72, -3.04] & 7.71e+08\\
         RRC vs. [ProgRRC and FRC] (C2) & 0.96 & [0.62, 1.31] & 9.91e+03 \\
         ProgRRC vs. FRC (C3) & 0.21 & [-0.18,0.60] & 0.178 \\
         \bottomrule  
    \end{tabular}
    \caption{Bayesian Logistic regression model estimates and Bayes Factors for the human experiment.}
    \label{tab:emprical-estimates-bfs}
\end{table}

\subsection{Statistical inferences for predicted data}\label{appendix:statistical-inference-predicted}

From the posteriors of the Bayesian models, we computed 95\% credible intervals for P(RRC $\mid$ prime, target) for each prime condition for the human data, and for each of our model types. If the credible intervals for predicted priming effect from a model do not overlap with the empirical priming effects, we infer that the model cannot account for human behavior. Such an inference is valid because credible intervals, unlike the frequentist confidence intervals, reflect our confidence about the distribution of the actual effects (so 95\% credible interval means that we are 95\% sure that the true effect falls within this interval). 

\section{Regular expressions to detect passive responses in the human experiment}\label{appendix:regex}

We used a three step process to detect passive responses in the human experiment. First, we started with the following regular expression: 
\begin{verbatim}
    ^(\\w+\\s+){3}by
\end{verbatim}
This expression looks for sentences in which the fourth word of the sentence is ``by''  --- all of our target prefixes had only three words (Determiner Noun Verb). 

Next we used the following regular expression to detect completions where the fourth word is ``by'', but the completion is not passive: 
\begin{verbatim}
     by \\w+(\\s+\\w+){0,1}(\\.)*$
\end{verbatim}
This expression returns TRUE if the word ``by'' is followed by just one or two words such as ``The thief chased by the dog'' or ``the thief chased by me''.

Finally, we tagged completions as being passive RRC completion if they matched the first expression and not the second. 
 \newpage

 \section{Limitations} \label{appendix:limitations}
Here we discuss some of the simplifying design decisions we made in SPAWN as a starting point, and their limitations. 

\paragraph{Storing discarded categories}
As discussed in \S~\ref{sec:reanalysis}, when the parser is regressing to some previous word $w_z$, it discards all of the categories retrieved from $w_z \dots, w_{i-1}, w_i$. In the current implementation, SPAWN stores all instances of the discarded categories and uses this to compute inhibition. While storing all instances of the discarded categories is convenient, it is not cognitively plausible. Future work can examine other ways of computing inhibition that relies on summaries of discarded categories, instead of storing all of the instances, and investigate if using summaries results in different priming behavior. 

\paragraph{Constraining partial parse states}
With our Whiz-Deletion grammar and our current implementation of null element prediction, the model could parse the partial sentence ``The cat examined'' and end up with an ungrammatical partial parse --- i.e., a parse that cannot result in a grammatical continuation --- as illustrated below. 

\begin{table}[h!]
    \centering
     \resizebox{0.9\textwidth}{!}{
    \begin{tabular}{lllll}
    \toprule
        \textbf{Old Parse state} & \textbf{Word} & \textbf{Retrieved category} & \textbf{Rule} & \textbf{New parse state}\\
        \midrule
        NULL & the & DP/NP & Initialize & DP/NP \\ 
        DP/NP & cat & NP/CP & Forward Harmonic Composition & DP/CP \\
        DP/CP & \color{red}{NULL\textsubscript{wh}} & CP/(TP$\backslash$DP) & Forward Harmonic Composition & DP/(TP$\backslash$DP) \\
        DP/VoiceP & examined & (TP$\backslash$DP)/DP & Forward Harmonic Composition & DP/DP \\
        \bottomrule
    \end{tabular}
    }
    \label{tab:ccg-analysis-PP-RRC}
\end{table}

This is not a problem in full sentences because this parse state is inconsistent with later words in the sentence, and the model will be forced to re-analyze. However, since our partial target prompts have no additional words, the model could end up with an ungrammatical parse, which is something we assumed would not happen with our human participants. Therefore, we constrained the model such that if it generated a partial state that was not DP/PP or TP/DP, it would be forced to reanalyze. While this is a convenient method to ensure that the model does not end up with an ungrammatical parse, it is unclear if this method accurately models how humans process the partial prompt. Future work can state more explicitly how humans parse the partial sentence such that they are always able to generate grammatical continuations, and then implement this in SPAWN.

\end{document}